\documentclass{article}
\usepackage{spconf,amsmath,graphicx}
\usepackage{amsfonts}
\usepackage{amssymb}
\usepackage{amsthm,amsmath}
\usepackage{mathrsfs}
\usepackage{indentfirst}
\usepackage{multirow}
\usepackage[colorlinks=true,citecolor=blue]{hyperref}
\title{A Multi-Stage Duplex Fusion ConvNet \\ for Aerial Scene Classification}
\name{Jingjun Yi, Beichen Zhou}%*\thanks{*Corresponding author: q\_bi@whu.edu.cn  \qquad \qquad \qquad \qquad \qquad \qquad   \linebreak This research is supported by the National Key Research and Development Program of China (No.2017YFB0503600) and the National Key Research and Development Program of China (No. 2016YFB0502600).}}
\address{School of Remote Sensing and Information Engineering, Wuhan University, China}
\begin{document}
%\ninept
%
\maketitle
\begin{abstract}
Existing deep learning based methods effectively prompt the performance of aerial scene classification. However, due to the large amount of parameters and computational cost, it is rather difficult to apply these methods to multiple real-time remote sensing applications such as on-board data preception on drones and satellites.
In this paper, we address this task by developing a light-weight ConvNet named \textit{multi-stage duplex fusion network} (MSDF-Net). The key idea is to use parameters as little as possible while obtaining as strong as possible scene representation capability. To this end, a residual-dense duplex fusion strategy is developed to enhance the feature propagation while re-using parameters as much as possible, and is realized by our \textit{duplex fusion block} (DFblock).
Specifically, our MSDF-Net consists of multi-stage structures with DFblock. Moreover, duplex semantic aggregation (DSA) module is developed to mine the remote sensing scene information from extracted convolutional features, which also contains two parallel branches for semantic description.
Extensive experiments are conducted on three widely-used aerial scene classification benchmarks, and reflect that our MSDF-Net can achieve a competitive performance against the recent state-of-art while reducing up to 80\% parameter numbers. Particularly, an accuracy of 92.96$\%$ is achieved on AID with only 0.49M parameters.  
\end{abstract}
\section{Introduction}
\label{sec1}

In recent years, with the rapid development of aerial sensors such as satellites, abundant aerial images have become accessible. Consequently, the computer vision community also pays increasingly attention on the application of aerial images, such as aerial scene classification \cite{Bi2019APDC,Bi2021MS}, land-cover reconstruction \cite{ratajczak2019automatic}, aerial object detection \cite{Xia2017DOTA,Ding2019Learning}. Aerial scene classification is the main task of aerial image interpretation, which aims to label semantic information of large-scale surface efficiently and automatically. This task is not trivial compared with generic ground scene classification, as the object distribution in aerial image is usually more complicated than ground scenes due to the large-scale bird view \cite{Bi2020A}.

Although great modification for aerial images on top of deep learning models have shown extraordinary talent in aerial scene classification, such models are usually rather heavy in terms of parameter numbers and computational cost. Unfortunately, real-time remote sensing applications such as on-board aerial image processing and preception from drones and satellites is also crucial, and the tremendous number of model parameters and the long inference time are unacceptable. Hence, light-weight ConvNets, which can effectively extract image features with only a small amount of parameters, are highly preferred for such applications \cite{Bi2020A}. 

\begin{figure}[t]
    \centering %插入的图片居中表示
	\includegraphics[width=3.6in]{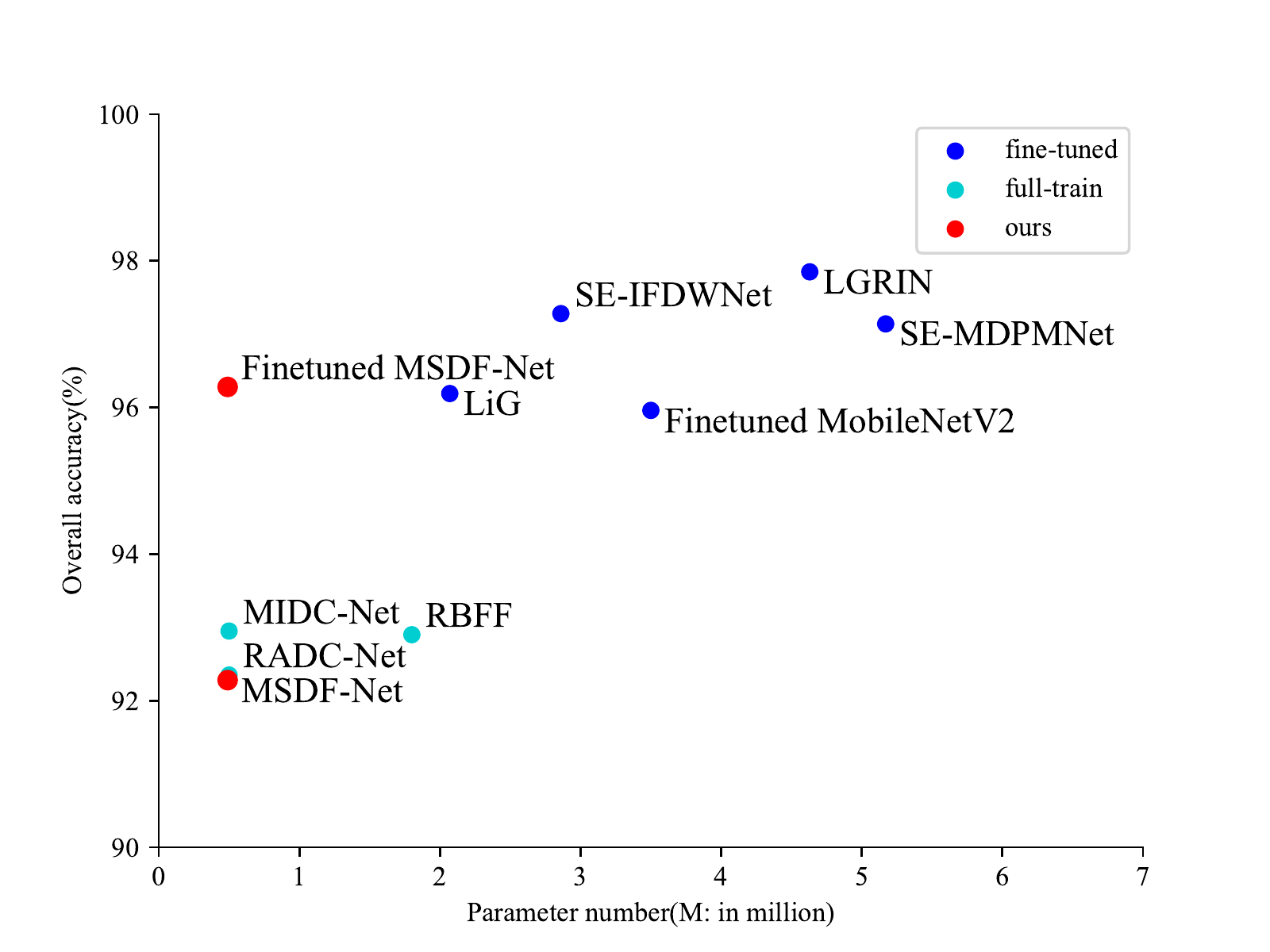}
	\vspace{-0.5cm}
	%插入的图，包括JPG,PNG,PDF,EPS等，放在源文件目录下
	\caption{Comparison of overall accuracy and model parameters (in million) between our proposed MSDF-Net and other present lightweight network for aerial scene classification.}  %图片的名称
	\vspace{-0.5cm}
	\label{fig1}   %标签，用作引用
\end{figure}

Presently, regarding light-weight ConvNets for aerial scenes, two major paradigms for light-weight network are \textit{adding modules} and \textit{new architecture} respectively. For \textit{adding modules}, current methods such as SE-MDPMNet \cite{zhang2019lightweight} and SE-IFDWNet \cite{bai2021lightweight} leverage some ingenious extra structures to enhance the feature representation from the original light-weight CNNs \cite{li2020automated}. Meanwhile, for \textit{new architecture}, high performance can also be obtained. Specially, for RADC-Net \cite{bi2020radc} and MIDC-Net \cite{bi2020multiple}, the effective structure design of convolutional layers enables them to use fewer parameters to achieve better results. However, adding modules directly on the end of CNN will lead to more computation cost to some extent. Therefore, the solution with practicability, universality and less computation is still preferred.

In this paper, we propose a light-weight ConvNet named MSDF-Net under the guidance of a residual-dense duplex fusion strategy. Specifically, in each duplex fusion block (DFblock), two commonly-used feature reuse solutions, namely residual connection and dense connection, are combined with multiple steps. To achieve this, apart from the residual fusion at the beginning and end of the block, the dense-connection structures are also cascaded with residual connections in dual branches. Notably, compared with previous methods, our model is more parameter-efficient and achieves higher accuracy on the benchmarks (see Fig.~\ref{fig1}~for intuitive comparison). 
Our contribution can be summarized as below.

(1) We propose a light-weight ConvNet named MSDF-Net for aerial scene classification. Notably, by combining use of multiple feature reuse and propagation strategies, it can achieve a competitive performance against the state-of-the-art while saving up to 80\% parameter numbers against prior art. 

(2) We propose a residual-dense duplex fusion strategy to extract multi-stage features by repetitively taking the advantage of the mixed use of residual connection and dense connection. We realize this strategy by designing duplex fusion block (DFblock), where we separate input feature channels into two parts to implement residual connection and dense connection respectively.

(3) We propose a duplex semantic aggregation (DSA) module, which is utilized as an enhanced semantic classifier and fuses the semantic probability distribution from duplex branches.   

\section{Related Work}
\label{sec2}
\subsection{Light-weight Networks for Aerial Scenes}
\label{ssec2.1}

The core of light-weight network is to optimize the network from both volume and speed on the premise of maintaining accuracy as much as possible. Lightweight networks like SquezeNet \cite{iandola2016squeezenet}, ShuffleNet \cite{zhang2018shufflenet} and MobileNet \cite{howard2017mobilenets} are widely used in real-time scene recognition and mobile deployment because of their low computational cost. The above networks do have a small amount of parameters and can achieve real-time inference, but the simplified network structure also leads to unrepresentative low-level features. At the same time, it will also lead to inadaptability when doing transfer learning or fine-tuning, where the feature distribution of two datasets may be widely divergent.

Presently, proposed lightweight networks for aerial scene can be divided into two categories, that is, \textit{adding modules} and \textit{new architecture}. For the first category, pre-trained models will be used as initial weights. Lin \textit{et al.} \cite{bai2021lightweight} proposed a lightweight multi-scale depthwide network with efficient spatial pyramid attention which use MobileNetV2 as a backbone. Md \textit{et al.} \cite{arefeen2021lightweight} proposed a layer selection strategy named ReLU-Based Feature Fusion which stacks features extracted from MobileNetV2. For the second category, it directly trains a new self-designed network structure on aerial image datasets. Xu \textit{et al.} \cite{xu2021lightweight} proposed the lie group regional influence network. Bi \textit{et al.} proposed RADC-Net \cite{bi2020radc} and MIDC-Net \cite{bi2020multiple} with adept self-designed structures.

Both categories have advantages and disadvantages. The premise of \textit{adding modules} is to have pre-trained models, which makes this method not flexible enough. At the same time, the additional convolution layers will increase the amount of parameters. Using \textit{new architecture} has no requirement for pre-trained models, which makes its application has stronger transferability.%but they can often ahieve higher accuracy in aerial scene.Training a new network has no requirements for pre-trained models, which makes its application havs stronger potentially transferability. Meanwhile, it is possible to have fewer parameters and simpler network structure. But the disadvantage is that accuracy of full training method usually can not reach the level of the previous method. 

\subsection{Residual and Dense Connection Structure}
\label{sec2.2}

In residual connection structure \cite{he2016deep}, residual block was widely used, which ensures feature representation capability. Later, dense connection structure \cite{huang2017densely} was proposed to further improve the information flow between convolution layers, as the dense block stacks features on the input features to better integrate new features with advanced features.

At present, many existing network structures are inspired by further exploiting these two structures. For our task, aerial scene classification with light-weight ConvNets, MIDC-Net proposed by Bi \textit{et al.} utilizes simplified dense connection structure with much fewer layers \cite{bi2020multiple}. In RADC-Net \cite{bi2020radc}, residual connection structure combined with spatial attention is stacked together with dense block. However, simply stacking the networks composed of these two structures can lead to the loss of original feature to a certain extent.

\section{Methodology}
\label{sec3}

\subsection{Framework Overview}
\label{sec3.1}
\begin{figure*}[t]
    \centering %插入的图片居中表示
	\includegraphics[height=3.4in]{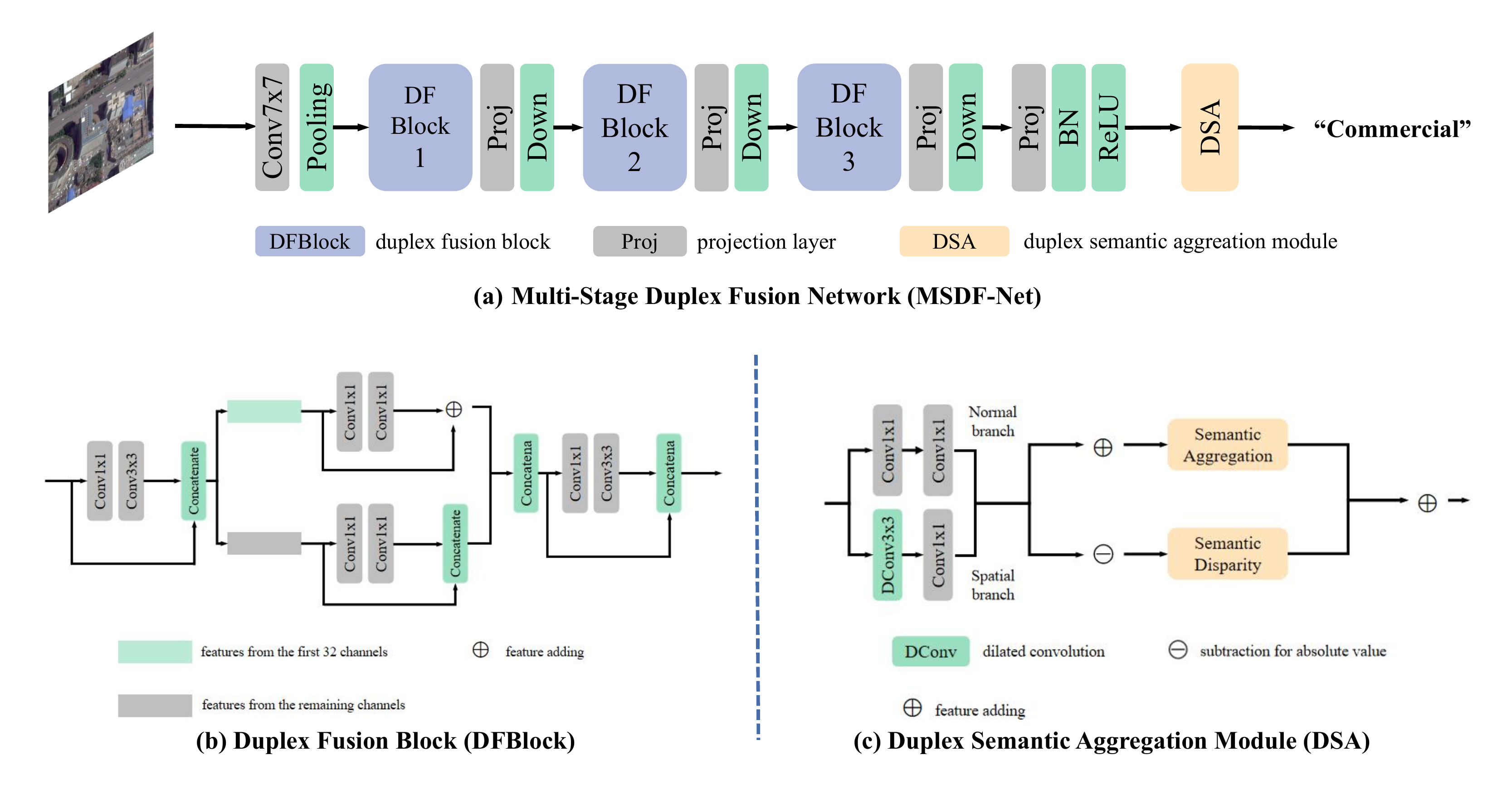}
	%\vspace{-0.5cm}
	%插入的图，包括JPG,PNG,PDF,EPS等，放在源文件目录下
	\caption{Illustration of our proposed multi-stage duplex fusion network (MSDF-Net) for aerial scene classification.}  %图片的名称
	\vspace{-0.5cm}
	\label{fig2}   %标签，用作引用
\end{figure*}

Explicit demonstration of our proposed MSDF-Net is given in Fig.~\ref{fig2}. After pre-processing, the initial convolutional feature was fed into DFblocks (Sec.~\ref{sec3.2}) to generate new features. Then, projection layers (Sec.~\ref{sec3.3}) are designed to represent the semantic differences between features of each DFblocks, so as to integrate features. After projection layer, average pooling is utilized to downscale features. Finally, the feature maps are processed by a classifier named duplex semantic aggregation module (Sec.~\ref{sec3.4}).

\subsection{Duplex Fusion Block}
\label{sec3.2}
The effective combination of residual and dense features can make the features more representative, which is of great significance for aerial scene classification. 

As shown in Fig.~\ref{fig2}, given a processed feature $X$ $\in$ $\mathbb{R}^{C \times H \times W}$ and a parameter $G$ which indicates the increase of feature maps, we feed it to a $1 \times 1$ convolution layer (denoted as $conv1_{1}$) and a $3 \times 3$ convolution layer (denoted as $conv3_{1}$) respectively, to generate initial dense feature $X_{1}$ $\in$ $\mathbb{R}^{G \times H \times W}$, presented as  $$X_{1}=conv3_{1}(conv1_{1}(X)).\eqno{(1)}$$

In order to calculate residual and dense connection respectively, we first concatenate $X$ and $X_{1}$ to get the intermediate feature $X_{2}$ $\in$ $\mathbb{R}^{(G + C) \times H \times W}$, and then split it into two parts $X_{3}$ $\in$ $\mathbb{R}^{G \times H \times W}$, $X_{4}$ $\in$ $\mathbb{R}^{C \times H \times W}$. For $X_{3}$, residual features are calculated from two $1 \times 1$ convolution layers (denoted as $conv1_{2}$ and $conv1_{3}$ respectively) and added to $X_{3}$ to obtain $X_{r}$ $\in$ $\mathbb{R}^{G \times H \times W}$. For $X_{4}$, dense features are given by two $1 \times 1$ convolution layers (denoted as $conv1_{4}$ and $conv1_{5}$ respectively) and concatenated (denoted as $concat$) to $X_{3}$ to obtain $X_{d}$ $\in$ $\mathbb{R}^{(G + C) \times H \times W}$, presented as
$$X_{r}=X_3 + conv1_{3}(conv1_{2}(X_3))).\eqno{(2)}$$
$$X_{d}=concat(X_4, conv1_{5}(conv1_{4}(X_4)).\eqno{(3)}$$

To fuse residual feature $X_{r}$ and dense feature $X_{d}$, we calculate a non-linear mapping of concatenated feature, and take it as output $X_{out}$ $\in$ $\mathbb{R}^{(G + 3C) \times H \times W}$. Significantly, through above operations we extract both residual and dense features at the same time, then the combination of non-linear mapping enables our MSDF-Net to learn richer semantic information and spatial details.  

\subsection{Projection Layers}
\label{sec3.3}
After the multi-layer convolution of DFblock, there is a lack of connection structure between the duplex fusion feature and the feature after down-sampling. In order to solve this problem, we design a projection layer to fill this gap. Our projection layer is composed of batch normalization (BN), Rectified Linear Units (ReLU), $1 \times 1$ convolution layer and outputs features of the same size as the input features. With projection layers act as bridge in our multi-stage training structure, the simple convolution processing ensures residual and dense features to be propagated more smoothly.

\subsection{Duplex Semantic Aggregation Module}
\label{sec3.4}
Feature maps are processed by our duplex semantic aggregation (DSA) module to generate the semantic prediction, which also contains duplex branches to exploit both semantic prediction and its difference from both branches. Take final features as $F$, it will be disposed respectively by semantic branch convolutional layers and spatial semantic branch convolutional layers. Normal semantic branch which composed of two $1 \times 1$ conv aims to extract semantic features $F_{1}$ at the general level, and for spatial semantic branch, one dilated convolution layer and one $1 \times 1$ conv was utilized to obtain spatial attention based semantic features $F_{2}$.
To achieve duplex semantic features aggregation, we reconcile the output of normal semantic branch $F_{1}$ and spatial semantic branch $F_{2}$ by summing them up. And final output $S_{agg}$ is presented as:
$$S_{agg} = F_{1} + F_{2} \eqno{(4)}$$

\begin{table*}[!t]\scriptsize
    \centering
    %\begin{threeparttable}
    \caption{Comparison of our MSDF-Net and other SOTA approaches, fine-tuning methods are marked with $*$, and direct-training methods are marked with $\dagger$. Metrics of accuracy are presented in \% and are in the form of \textit{'mean accuracy $\pm$ standard deviation'}; following the evaluation protocols \cite{Yi2013Geographic,Xia2017AID,Gong2017Remote}; The number of parameters is measured in millions (M); the model size is measured in megabytes (MB).}
    \begin{tabular}{ccccccccc} 
    \hline
    Method & \multicolumn{2}{c}{UCM} &	\multicolumn{2}{c}{AID} &	\multicolumn{2}{c}{NWPU} &  Parameters &    Model size\\ 
    \cline{2-7}
    ~ & 50\% & 80\% &20\% & 50\% & 10\% & 20\%\\
    %\hline
    %AlexNet \cite{Xia2017AID} & 93.98$\pm$0.67 & 95.02$\pm$0.81 & 86.86$\pm$0.47 & 89.53$\pm$0.31 & 76.69$\pm$0.21 & 79.85$\pm$0.13\\
    %VGGNet-16 \cite{Xia2017AID} & 94.14$\pm$0.69 & 95.21$\pm$1.20 & 86.59$\pm$0.29 & 89.64$\pm$0.36 & 76.47$\pm$0.18 & 79.79$\pm$0.15\\
    %GoogLeNet \cite{Xia2017AID} & 92.70$\pm$0.60 & 94.31$\pm$0.89 & 83.44$\pm$0.40 & 86.39$\pm$0.55 & 76.19$\pm$0.38 & 78.48$\pm$0.26\\
    \hline
    Fine-tuned MobileNetV2$^{*}$ \cite{zhang2019lightweight} & ---- & ----
    & 94.13$\pm$0.28 & 95.96$\pm$0.27 
    & 90.16$\pm$0.12 & 93.00$\pm$0.18
    & 3.5 & 39.7\\
    SE-MDPMNet$^{*}$ \cite{zhang2019lightweight} & ---- & ---- 
    & 94.68$\pm$0.17 & 97.14$\pm$0.15  
    & 91.80$\pm$0.07 & 94.11$\pm$0.03
    & 5.17 & ----\\
    SE-IFDWNet$^{*}$ \cite{bai2021lightweight}& ---- & ---- 
    & 94.70$\pm$0.12 & 97.28$\pm$0.22  
    & 91.88$\pm$0.15 & 94.14$\pm$0.11
    & 2.86 & ----\\
    %\hline
    %\hline
    %\hline
    RBFF$^{*}$ \cite{arefeen2021lightweight} & 94.69$\pm$0.34 & 96.40$\pm$0.99 
    & 90.94$\pm$0.26 & 93.73$\pm$0.23 
    & 84.59$\pm$0.24 & 87.53$\pm$0.19
    & 1.46 & 6.28\\
    %\hline
    %FV \cite{Li2017Integrating} & ---- & 98.57$\pm$0.34& ---- & ----& ---- & ----\\
    %\hline
    %ESPA-MSDWNet \cite{bai2021lightweight} & ---- & ----
    %& \textbf{94.24$\pm$0.10} & \textbf{98.46$\pm$0.04} 
    %& \textbf{91.08$\pm$0.13} & \textbf{95.28$\pm$0.08}\\
    \textbf{Fine-tuned MSDF-Net$^{*}$} (ours) & \textbf{97.03$\pm$0.58} & \textbf{98.46$\pm$0.59} 
    & \textbf{93.01$\pm$0.25} & \textbf{96.20$\pm$0.16} 
    & \textbf{86.27$\pm$0.12} & \textbf{90.88$\pm$0.08}
    & \textbf{0.49} & \textbf{5.92}\\
    \hline
    MIDC-Net\_S$^{\dagger}$ \cite{bi2020multiple} & 94.93$\pm$0.51 & 97.00$\pm$0.49
    & 88.26$\pm$0.43 & 92.53$\pm$0.18
    & 85.59$\pm$0.26 & 87.32$\pm$0.17
    & 0.5 & 9.9\\
    RADC-Net$^{\dagger}$ \cite{bi2020radc} & 94.79$\pm$0.42 & 97.05$\pm$0.48
    & 88.12$\pm$0.43 & 92.35$\pm$0.19 
    & \textbf{85.72$\pm$0.25} & 87.63$\pm$0.28
    & 0.5 & 9.0\\
    %\hline
    \textbf{MSDF-Net$^{\dagger}$} (ours) & \textbf{96.90$\pm$0.24} & \textbf{97.38$\pm$0.23} 
    %& 89.36$\pm$0.18 & 92.28$\pm$0.26 
    %& 80.14$\pm$0.14 & 84.67$\pm$0.12\\
    & \textbf{89.92$\pm$0.18} & \textbf{92.96$\pm$0.20} 
    & 85.14$\pm$0.16 & \textbf{87.67$\pm$0.12}
    & \textbf{0.49} & \textbf{5.25}\\
    \hline
    \end{tabular} 
     \label{tab1}
    %\end{threeparttable}
\end{table*}

During training procedure, in order to further aggregate two semantic branches and gradually reduce the semantic difference between the final features of two branches, we design a dual loss training framework for MSDF-Net. In this framework, the original single probability output is replaced by dual loss $L_{dual}$ which is a weighted sum of normal loss $L_{N}$ and disparity loss $L_{R}$, presented as:
$$L_{N} = crossentropy(softmax(S_{agg})) . \eqno{(5)}$$
$$L_{R} = crossentropy(softmax(|F_{1} - F_{2}|)) . \eqno{(6)}$$
$$L_{dual} = L_{N} + \alpha * L_{R}. \eqno{(7)}$$

The value of $\alpha$ in the above formula is 5e-4. And for dual loss, we utilize cross-entropy cost function(denoted as $crossentropy$) as loss function.

\section{Experiment and analysis}
\label{sec4}

\subsection{Benchmark, Evaluation Metrics and Settings}

\textbf{Benchmark.} We verify our proposed method on three widely-used aerial scene classification benchmarks, namely UC Merced (UCM) \cite{Yi2013Geographic}, Aerial Image Dataset (AID) \cite{Xia2017AID} and Northwestern Polytechnical University (NWPU) \cite{Gong2017Remote}. 

\textbf{Evaluation Metrics.} All these benchmarks are divided into training and testing set to conduct comparative experiments \cite{Yi2013Geographic,Xia2017AID,Gong2017Remote}. Specially, For UCM, AID and NWPU, the training ratios are 50\% \& 80\% \cite{Yi2013Geographic}, 20\% \& 50\% \cite{Xia2017AID} and 10\% \& 20\% \cite{Gong2017Remote} respectively.

\textbf{Hyper-parameter settings.} %Tensorflow was utilized to implement our proposed RDFNet. 
Following former works, the samples were processed by rotation, light adjustment and flip as data augmentation, and dropout rate is set to 0.2. We utilize adam optimizer to optimize our network. The initial learning rate was set to 0.001, and decay to half the original every 30 epochs until 120 epochs are finished. We train and test our model on a PC with a NVIDIA Geforce GTX1650 GPU. %For dual rest loss, we set $\alpha$ as 5e-4.%For optimizer, we use Adam optimizer to train our framework. The parameters and settings of our training procedure are shown in detail below.   

% \begin{table}[!t]  
%     \centering
%     \begin{threeparttable}
%     \caption{Comparison of number of parameters of our RDFNet and other methods on AID dataset.}
%     %\vspace{0.5em}
%     \begin{tabular}{ccc}
%     %\vspace{0.5em}
%     \hline
%     %\cline{2-3}
%     %\hline
%     %Finetuned-MobileNetV2 & 3.5 & 91.72$\pm$0.17 \\
%     %SE-MDPMNet & 5.17 & 91.72$\pm$0.17 \\
%     %RBFF & 13.28 & 91.72$\pm$0.17 \\
%     %ESPA-MSDWNet & 2.4 & 91.72$\pm$0.17 \\
%     %\hline
%     Method & Parameters(M) & Model size(MB) \\
%     \hline
%     MIDC-Net\_S & 0.5 & 9.9 \\
%     RADC-Net & 0.5 & 9.0 \\
%     \textbf{RDFNet} & \textbf{0.49} & \textbf{5.25} \\
%     \hline
%     \end{tabular} 
%      \label{tab2}
%     \end{threeparttable}
% \end{table}

\subsection{Comparison with the State-of-the-art Methods}
All experimental results of our MSDF-Net and existing state-of-the-art (SOTA) methods on these three datasets under six different experiments are listed in Table.~\ref{tab1}. The number of parameters and model size is also shown in Table.~\ref{tab1}. Clearly, our proposed MSDF-Net pushes the frontiers as it achieves a competitive performance against existing state-of-the-art while only need much fewer parameter number and model size.

\begin{table}[!t]  
    \centering
    %\begin{threeparttable}
    \caption{Ablation study of our MSDF-Net on AID dataset. Proj: projection layer; DSA: duplex semantic aggregation; Metrics are presented in \% and are in the form of \textit{'mean accuracy $\pm$ standard deviation'} following the evaluation protocols in \cite{Xia2017AID}.}
    \begin{tabular}{ccc} 
    \hline
    %\cline{2-3}
    ~ & 20\% & 50\% \\
    \hline
    MSDF & 88.63$\pm$0.26 & 91.72$\pm$0.17 \\
    MSDF+proj & 89.36$\pm$0.18 & 92.28$\pm$0.26 \\
    \textbf{MSDF+Proj+DSA} & \textbf{89.92$\pm$0.18} & \textbf{92.96$\pm$0.20} \\
    \hline
    \end{tabular} 
     \label{tab2}
    %\end{threeparttable}
\end{table}

\subsection{Ablation Studies}

In our MSDF-Net, pooling layers and DFBlock are connected by \textit{projection layers} (Proj), while \textit{duplex semantic aggregation module} (DSA module) is utilized as a enhanced classifier. To find out if projection layers and our proposed duplex semantic fusion module help smooth feature connection and refine differential features. Experiments are conducted in our ablation studies on AID benchmark, and the results are listed in Table~\ref{tab2}. It can be observed that all these components contribute positively to scene representation learning.

\section{Conclusion}
\label{sec5}
We propose a light-weight MSDF-Net for aerial scene classification. The DFblocks are utilized to integrate residual and dense connections under the gudiance of our proposed duplex fusion strategy. Then, duplex semantic fusion module is used as an enhanced classifier for final feature maps. Extensive Experiments on three aerial scene classification datasets verify that our proposed MSDF-Net can achieve comparable performance with much fewer parameters than several state-of-the-art.

% Below is an example of how to insert images. Delete the ``\vspace'' line,
% uncomment the preceding line ``\centerline...'' and replace ``imageX.ps''
% with a suitable PostScript file name.
% -------------------------------------------------------------------------

\iffalse

\begin{figure}[htb]

\begin{minipage}[b]{1.0\linewidth}
  \centering
  \centerline{\includegraphics[width=8.5cm]{image1}}
%  \vspace{2.0cm}
  \centerline{(a) Result 1}\medskip
\end{minipage}
%
\begin{minipage}[b]{.48\linewidth}
  \centering
  \centerline{\includegraphics[width=4.0cm]{image3}}
%  \vspace{1.5cm}
  \centerline{(b) Results 3}\medskip
\end{minipage}
\hfill
\begin{minipage}[b]{0.48\linewidth}
  \centering
  \centerline{\includegraphics[width=4.0cm]{image4}}
%  \vspace{1.5cm}
  \centerline{(c) Result 4}\medskip
\end{minipage}
%
\caption{Example of placing a figure with experimental results.}
\label{fig:res}
%
\end{figure}
\fi

%\vfill
%\pagebreak

\vfill\pagebreak

\bibliographystyle{IEEEbib}
\bibliography{refs}

\end{document}